\documentclass{article}

\PassOptionsToPackage{numbers, compress}{natbib}
\usepackage[final]{neurips_2024}

\usepackage[utf8]{inputenc} 
\usepackage[T1]{fontenc}    
\usepackage[hidelinks]{hyperref}       
\usepackage{url}            
\usepackage{booktabs}       
\usepackage{amsfonts}       
\usepackage{nicefrac}       
\usepackage{microtype}      
\usepackage{xcolor}         
\usepackage{multirow}
\usepackage{amsmath}
\usepackage{amssymb}
\usepackage{mathtools}
\usepackage{amsthm}
\usepackage{tikz}
\usepackage{xspace}
\usepackage{enumitem}
\usepackage{lipsum}
\usepackage[font=small,labelfont=bf]{caption}
\setlist[itemize]{leftmargin=*}
\usepackage{wrapfig}

\newcommand{\acro}[1]{\textsc{#1}\xspace}
\newcommand{\ginn}{\textsc{g}rad\textsc{inn}}
\newcommand{\ginns}{\ginn{s}}
\newcommand{\nn}{\acro{nn}}
\newcommand{\nns}{\acro{nn}{s}\xspace}
\newcommand{\pinn}{\acro{pinn}}
\newcommand{\pinns}{\acro{pinn}{s}\xspace}
\newcommand{\upinn}{\acro{upinn}}
\newcommand{\dhpm}{\acro{dhpm}}
\newcommand{\rmse}{\acro{rmse}}

\newcommand{\de}{\acro{de}}
\newcommand{\pde}{\acro{pde}}
\newcommand{\ode}{\acro{ode}}
\newcommand{\friedman}{\acro{friedman}}
\newcommand{\stokes}{\acro{stokes}}
\newcommand{\burgers}{\acro{burger}}
\newcommand{\lv}{\acro{lv}}

\newcommand{\snn}{s-\nn}

\newcommand{\vecx}{\boldsymbol{x}}
\newcommand{\xdim}{d}

\newcommand{\sol}{u}
\newcommand{\netsol}{\MakeUppercase{\sol}}

\newcommand{\netauxp}{N}
\newcommand{\diffoperator}{\mathcal{\netauxp}}

\newcommand{\netauxg}{F}

\newcommand{\netauxgso}{G}
\newcommand{\dataTp}{\mathcal{D}}
\newcommand{\dataCp}{\mathcal{B}}
\newcommand{\dataTg}{\dataTp_{\acro{train}}}
\newcommand{\dataCg}{\mathcal{C}}

\newcommand{\indexT}{n}
\newcommand{\indexTest}{h}
\newcommand{\ndataTp}{\MakeUppercase{\indexT}_{\netsol}}
\newcommand{\ndataTg}{\MakeUppercase{\indexT}_{\netsol}}

\newcommand{\indexC}{m}
\newcommand{\ndataCp}{\MakeUppercase{\indexC}}
\newcommand{\ndataCg}{\MakeUppercase{\indexC}}

\newcommand{\udim}{{d_o}}
\newcommand{\indexu}{{k}}

\newcommand{\ntest}{{N_\acro{test}}}

\title{GradINN: Gradient Informed Neural Network}

\author{%
Filippo Aglietti$^{1, 2}$ \quad Francesco {Della Santa}$^{3}$ \quad Andrea Piano$^1$ \quad Virginia Aglietti$^4$\\
\quad \\
$^1$Energy Department, Politecnico di Torino, Torino, Italy \\ $^2$Dumarey Automotive Italia S.p.A. \\ $^3$Department of Mathematical Sciences, Politecnico di Torino, Torino, Italy \\ $^4$ Independent Researcher\\
\quad \\
\texttt{filippo.aglietti@studenti.polito.it}\\
\texttt{francesco.dellasanta@polito.it}\\
\texttt{andrea.piano@polito.it}\\
\texttt{agliettivirginia@gmail.com}
}

\begin{document}

\maketitle

\begin{abstract}
We propose Gradient Informed Neural Networks (\ginns), a methodology inspired by Physics Informed Neural Networks (\pinns) that can be used to efficiently approximate a wide range of physical systems for which the underlying governing equations are completely unknown or cannot be defined, a condition that is often met in complex engineering problems. 
\ginns~leverage \textit{prior beliefs} about a system's gradient to constrain the predicted function's gradient across all input dimensions.  
This is achieved using two neural networks: one modeling the target function and an auxiliary network expressing prior beliefs, e.g., smoothness. 
A customized loss function enables training the first network while enforcing gradient constraints derived from the auxiliary network. We demonstrate the advantages of \ginns, particularly in low-data regimes, on diverse problems spanning non-time-dependent systems (Friedman function, Stokes Flow) and time-dependent systems (Lotka-Volterra, Burger's equation). Experimental results showcase strong performance compared to standard neural networks and \pinn-like approaches across all tested scenarios.
\end{abstract}

\section{Introduction}
\label{intro}
In the field of computational physics, Neural Networks (\nns) have become an increasingly important tool for modeling complex physical systems that cannot be derived in closed form or for which the traditional empirical models fail to achieve the desired accuracy \cite{ORESKI2012303, en13205301}. Several studies have shown how \nns~are powerful function approximators, able to model a wide variety of large, complex, and highly non-linear systems with unprecedented computational efficiency when a large training dataset is available \cite{HORNIK1989359, Pinkus_1999, 10112753, JMLR:v24:22-0384}. However, in settings where data is limited or widely dispersed, \nns~face considerable difficulties. For instance, in the physical sciences, data is often obtained experimentally and is thus expensive and/or challenging to collect. In such scenarios, \nns~show a decreasing prediction performance and a higher probability to overfit to the training data compared to physics white/gray models. On the other hand, the latter suffer from lack of flexibility and expressivity (for instance 0/1-dimensional models) or can require high computational effort as Finite Element Method models \cite{reddy2013introduction}. 
In order to address these difficulties, Physics-Informed
Neural Networks (\pinns) emerged in recent years \cite{RAISSI2019686}. These models leverage prior knowledge, often in the form of known differential equations (\de), by embedding it directly into the training process. Specifically, they introduce an additional term in the loss function which represents the residual of the underlying \de evaluated on a set of so-called collocation points. This formulation increases robustness against flawed data, e.g., missing or noisy values, and offers physically consistent predictions, particularly in tasks requiring extrapolation \cite{Karniadakis}. While \pinns have been shown to perform well across a variety of applications \cite{RAISSI2019686, Karniadakis, 9743327, MAO2020112789, 10.1115/1.4050542, YU2022114823, cai2021physics, noguer2023physics} they require \textit{detailed prior knowledge} of the physical system and are thus not directly applicable when this is not available. 
Recently, \pinns have been extended to deal with various modified settings: systems characterized by \textit{partially unknown} underlying physics \cite{Raissi_2018, 10.5555/3618408.3619569}, unknown data measurement noise \cite{pilar2023physicsinformed}, gradients observations \cite{son2021sobolev} by using Sobolev training \cite{sobolev} and learning of 
symplectic gradients (Hamiltonian \nns,  \cite{hamiltonian, hamiltonian2}). 

However, there exists many physical phenomena where the underlying physics is \emph{entirely unknown} or too complex to be easily represented through \de{s}. In those cases it is only possible to rely only on data and \emph{prior beliefs}. 
\ginns, gradient informed neural networks, address these settings by differentiating, similarly to \pinns, a primary \nn~on a set of collocation points. However, differently from \pinns, \ginns~pair the primary network with an auxiliary \nn encoding prior belief and leading to an additional loss term which regularizes the predicted solution gradients.
As a result, \ginns~can effectively model a broader class of physical systems. In summary, our contributions are as follow:
\begin{itemize}
    \item We propose \ginn, a simple and efficient technique for training a \nn to accurately approximate a function in low data regimes. \ginn~constraints the \nn's gradient via an auxiliary network that encodes prior beliefs about the underlying system, and trains both networks via a customized loss that can be easily incorporated into any training pipeline.
    \item We test \ginn{s} on a general synthetic function (Friedman) and show how they outperform other \nns, with and without regularization, while being more robust to noisy data.
    \item We extensively test \ginn{s} in the context of physical systems. We show how they can be used to learn the solution of ordinary differential equations (\ode{s}, e.g. Lotka-Volterra system), partial differential equations (\pde{s}, e.g. Burger's equation) and systems for which time is not the primary driver of change (Stokes Flow) outperforming \nns across all examples.
\end{itemize}

\section{Preliminaries}\label{sec:preliminaries}
\textbf{Notation} We denote by $\sol(\cdot):\mathbb{R}^{\xdim}\rightarrow\mathbb{R}$ a function we aim to model with inputs given by $(t, \boldsymbol{\xi})$, where $t \in \mathbb{R}$ is time and $\boldsymbol{\xi}:= [\xi_1,\ldots ,\xi_{\xdim-1}]\in \mathbb{R}^{\xdim-1}$ is the vector of spatial variables. 
To avoid cluttering notation we denote $\sol(\cdot)$ by $\sol$ hereinafter. Let $u_{t}:= \partial u/\partial t$ be the partial derivative of $u$ with respect to $t$ and $\sol_{\xi_i}:= \partial u/\partial \xi_i$ be the partial derivative of $u$ with respect to the $i$-th spatial variable $\xi_i$. Moreover, we denote by $\nabla_{\boldsymbol{\xi}}$ the differential operator that returns the vector of partial derivatives of a function with respect to $\boldsymbol{\xi}$ and with $\nabla^2_{\boldsymbol{\xi}}$ the operator for computing the matrix of second-order partial derivatives with respect to $\boldsymbol{\xi}$. Therefore, $\nabla_{\boldsymbol{\xi}}\sol := [\sol_{\xi_1}, \dots, \sol_{\xi_{d-1}}] \in \mathbb{R}^{\xdim-1}$ and $\nabla^2_{\boldsymbol{\xi}}\sol := (u_{\xi_i \xi_j}) = (\partial^2 u / \partial \xi_i \partial \xi_j) \in \mathbb{R}^{(\xdim-1) \times (\xdim-1)}$.



As originally formulated in \citet{Raissi_2018}, \pinns study physical systems governed by a known \pde of the general form:

\begin{equation}
    \sol_t = \diffoperator(t, \boldsymbol{\xi}, \sol, \nabla_{\boldsymbol{\xi}}\sol, \nabla^2_{\boldsymbol{\xi}}\sol, \ldots),
    \label{eq:PINN_eq}
\end{equation}

where $\diffoperator$ represents a known potentially nonlinear differential operator that depends on $t$, on the spatial variables $\boldsymbol{\xi}$, on $\sol$ and on its derivatives with respect to $\boldsymbol{\xi}$.
\pinns approximate the solution $\sol$ via a \nn denoted by $\netsol(\cdot; \Theta):\mathbb{R}^{d}\rightarrow\mathbb{R}$ where \(\Theta_{\netsol}\) are the network parameters.
From Eq. \eqref{eq:PINN_eq} it is possible to define the function $f~\coloneqq~\sol_t-\diffoperator(t, \boldsymbol{\xi}, \sol, \nabla_{\boldsymbol{\xi}}\sol, \nabla^2_{\boldsymbol{\xi}}\sol, \ldots)$ and, by plugging in the network $\netsol$, write:
\begin{equation}
    f := \netsol_t(t,\boldsymbol{\xi}; \Theta_{\netsol}) - \diffoperator(t, \boldsymbol{\xi}, U, \nabla_{\boldsymbol{\xi}}U, \nabla^2_{\boldsymbol{\xi}}U, \ldots).
    \label{eq:PINN_eq_f_U}
\end{equation}
where the derivatives $\netsol_t, \nabla_{\boldsymbol{\xi}}U, \nabla^2_{\boldsymbol{\xi}}U, \dots$ of the network $\netsol$ are computed by automatic differentiation \cite{automatic_diff}. Given a training dataset $\dataTp = \{(t_\indexT, \boldsymbol{\xi}_\indexT), \sol_\indexT\}_{\indexT=1}^{\ndataTp}$, including the initial and boundary conditions on $\sol$, and a set of collocation points \(\dataCp = \{(t_{\indexC}, \boldsymbol{\xi}_{\indexC})\}_{\indexC=1}^{\ndataCp}\), the parameters \( \Theta_{\netsol}\) can be learned by minimizing the loss function \(\mathcal{L}\) defined as: 
\begin{align}
\mathcal{L}(\Theta_{\netsol}) &= \mathcal{L}_{\netsol}(\Theta_{\netsol})  + \mathcal{L}_{f}(\Theta_{\netsol}) \nonumber \\
& = 
\frac{1}{\ndataTp} \sum_{\indexT=1}^{\ndataTp} (\netsol(t_{\indexT}, \boldsymbol{\xi}_{\indexT};\Theta_{\netsol}) - \sol_\indexT)^2+ \frac{1}{\ndataCp} \sum_{\indexC=1}^{\ndataCp}  (f(t_{\indexC}, \boldsymbol{\xi}_{\indexC};\Theta_{\netsol}))^2.
\label{eq:PINN_eq_loss}
\end{align}
The term $\mathcal{L}_{\netsol}$ constraints the solution $\sol$ to fit the training data $\dataTp$ whereas $\mathcal{L}_f$ enforces the physics knowledge dictated by Eq. \eqref{eq:PINN_eq_f_U} across the set of collocation points $\dataCp$. The Deep Hidden Physics Models (\dhpm{s}) methodology proposed in \cite{Raissi_2018}, similarly to the Universal-\pinn (\upinn) approach of \cite{10.5555/3618408.3619569}, extends the \pinn formulation to deal with physics problems governed by \pde{s} (or \ode{s}) of the form given in Eq. \eqref{eq:PINN_eq} and partially unknown \(\diffoperator\). In particular, \citet{Raissi_2018} considers unknown functional form but known input variables for $\diffoperator$ while \citet{10.5555/3618408.3619569} also assumes partial prior knowledge of the parameters of the functional form. In settings where partial knowledge exists, an auxiliary network $\netauxp(\cdot;\Theta_{\netauxp}): \mathbb{R}^D \to \mathbb{R}$ is used to approximate \(\diffoperator\). Note that the inputs to $\netauxp$ (and thus the value of $D$) are problem specific and correspond to the inputs of $\diffoperator$ thus encoding partial knowledge of the system. By plugging $\netauxp$ into Eq. \eqref{eq:PINN_eq_f_U} we can write:
\begin{equation}
    f := \netsol_t(t,\boldsymbol{\xi}; \Theta_{\netsol}) - \netauxp(t, \boldsymbol{\xi}, \netsol, \nabla_{\boldsymbol{\xi}}\netsol, \nabla^2_{\boldsymbol{\xi}}\netsol, \ldots; \Theta_{\netauxp}).
    \label{eq:PINN_eq_f_upinn}
\end{equation}
In this case, \((\Theta_{\netsol}, \Theta_{\netauxp})\) are learned considering the same \(\mathcal{L}_{\netsol}\) and a modified version of \(\mathcal{L}_f\) in which \(f\) is given by Eq. \eqref{eq:PINN_eq_f_upinn}. 
The training dataset $\dataTp$ is then augmented to include, in addition to initial and boundary conditions, data points within the domain of $\sol$ so as to compensate for the lack of prior knowledge of the system governing \de{s}.  

In the next section, we illustrate a new approach that is partially inspired by this \pinns formulation, but that addresses cases where the form of Eq. \eqref{eq:PINN_eq} (and as a consequence the form of $\netauxp$) is fully unknown or cannot be defined thus only data and prior beliefs about the system are available. 

\section{\ginn}\label{sec:method}
The approach we propose in this paper, which we name Gradient Informed Neural Network (\ginn), targets systems of the form $\sol(\vecx) \in \mathbb{R}^\udim$, where $\vecx \in \mathbb{R}^\xdim$ represents all the dimensions of the problem making no distinction between temporal and spatial variables.
As done in Section \ref{sec:preliminaries}, we denote by $\nabla_{\vecx}$ and $\nabla^2_{\vecx}$ the differential operators for the first and second order derivatives with respect to all the dimensions of vector $\vecx$ respectively.\footnote{Note that this notation is adopted to distinguish the \nn differentiation with respect to $\vecx$ instead of the model's weights.}
We first introduce the methodology for uni-dimensional outputs ($\udim=1$) and then discuss the extension to multi-output systems ($\udim>1$).

\begin{figure}[t]
\centering
\includegraphics[width=0.9\columnwidth]{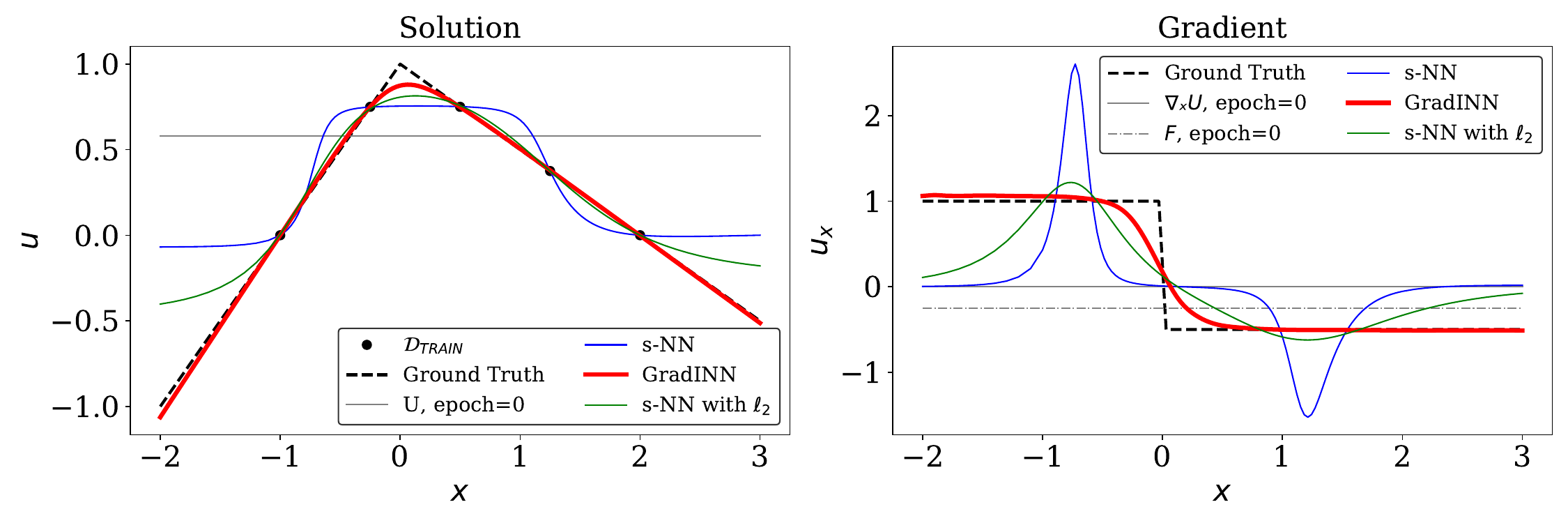}
\caption{Ground truth (black dashed lines) and predicted solution (left) and gradient (right) both at initialization (gray lines) and after training. Predictions for \ginn~are obtained with $\ndataCg=100$ collocation points uniformly distributed $[-2,3]$.
}
\label{LinSystem_solution}
\vskip -0.1in
\end{figure}

\ginn~deals with systems where the governing equation, i.e., Eq. \eqref{eq:PINN_eq}, is unknown or cannot be defined, and only general behaviour for $\nabla_{\vecx}\sol$ can be assumed. Similar to Eq.~\eqref{eq:PINN_eq_f_upinn},
\ginn~consists of two paired \nns, $\netsol(\cdot; \Theta_{\netsol}):\mathbb{R}^d\rightarrow\mathbb{R}$ and $\netauxg(\cdot; \Theta_{\netauxg}):\mathbb{R}^d\rightarrow\mathbb{R}^d$. $\netsol$ is used to approximate $\sol$ while $\netauxg$, taking $\vecx$ as input and returning the beliefs about $\nabla_{\vecx}\sol$ as output, is used to apply implicit constraints on $\nabla_{\vecx}\netsol$. 
In particular, the \textit{initialization} of $\netauxg$ encodes general \textit{prior} beliefs on the behavior of $\nabla_{\vecx}\sol$, e.g., smoothness, and, therefore, of $\sol$ itself. 
For instance, $\netauxg$ can be initialized to give constant outputs to encourage smoothness in the predicted solution given by $U$ (see below and Fig. \ref{LinSystem_solution} for a toy example). Given a training dataset $\dataTg = \{\vecx_\indexT, \sol_\indexT\}_{\indexT=1}^{\ndataTg}$ and a set of collocation points $\dataCg = \{\vecx_\indexC\}_{\indexC=1}^{\ndataCg}$, we train the paired \nns simultaneously by considering the following loss function:
\begin{equation}
\begin{aligned}
    \mathcal{L}(\Theta_U , \Theta_F ) &= \mathcal{L}_U(\Theta_U ) + \mathcal{L}_{F}(\Theta_F ; \Theta_U) = \\
    &= \mathcal{L}_U(\Theta_U) + \frac{1}{M}\sum_{\indexC=1}^{\ndataCg} \Vert F(\vecx_m;\Theta_F) - \nabla_{\vecx}\netsol(\vecx_m; \Theta_U) \Vert _2^2 = \\
    & = \mathcal{L}_U(\Theta_U) + \frac{1}{M}\sum_{\indexC=1}^{\ndataCg}\sum_{i=1}^d \left( F_i(\vecx_m;\Theta_F) - U_{x_i}(\vecx_m; \Theta_U) \right)^2\,,
\end{aligned}
\label{eq:GINN_eq_loss}
\end{equation}
where the first component $\mathcal{L}_{\netsol}(\Theta_{\netsol})$ is defined, similarly to Eq. \eqref{eq:PINN_eq_loss}, as
$\mathcal{L}_{\netsol}(\Theta_{\netsol}) = \frac{1}{\ndataTg} \sum_{\indexT=1}^{\ndataTg}\left(\netsol(\vecx_\indexT; \Theta_{\netsol}) - \sol_\indexT\right)^2$. Note how $\mathcal{L}_{\netsol}$ optimizes $\Theta_{\netsol}$ to have $U(\vecx;\Theta_{\netsol})\approx u(\vecx)$ over $\dataTg$. At the same time, $\mathcal{L}_{\netauxg}$ allows optimizing both $\Theta_{\netsol}$ and $\Theta_{\netauxg}$ such that $\netauxg(\vecx;\Theta_F)~-~\nabla_{\vecx}\netsol(\vecx;\Theta_U)~\approx~\boldsymbol{0}$ for $\vecx \in \dataCg$. In particular, by updating $\Theta_{\netsol}$, this loss term encourages $\netsol$ to have a gradient behavior similar to $\netauxg$. Simultaneously, it modifies $\Theta_{\netauxg}$ to bring $\netauxg$ closer to $\nabla_{\vecx}\netsol$, thereby relaxing the prior beliefs on $\dataCg$ using the gradient information derived from $\dataTg$ via $\mathcal{L}_{\netsol}$. As $\dataCg$ does not require access to the output values, it can be increased without the need to collect expensive experimental data but only accounting for the available computation resources or model efficiency considerations. Finally, analogously to Eq. \eqref{eq:PINN_eq_loss}, the proposed loss can be minimized using backpropagation and off-the-shelf optimizers thus being easily incorporated into any training pipeline. 

To further clarify the impact of the proposed loss (Eq. \eqref{eq:GINN_eq_loss}) and shed light of \ginn's working mechanism, we show a toy example where \ginn, a standard \nn\footnote{Here and in the rest of the paper, we call standard \nn a \nn trained only with $\mathcal{L}_{\netsol}$ on $\dataTg$.} (denoted by \snn~henceforth), and a \snn with $\ell_2$-regularisation are trained on a simple system $u:[-2,3]\rightarrow\mathbb{R}$, ($\xdim$=1, $\udim$=1), using $\ndataTg=5$ training points. Fig.~\ref{LinSystem_solution} shows the ground truth solution (left, black line) and gradient (right, black line) together with $\dataTg$ (black dots) and the predictions both at initialization (gray lines; note that the initialization of $\netsol$ is the same for \snn, \snn~with $\ell_2$ and \ginn) and after training. Notice how, initializing $\netsol$ and $\netauxg$ in a way that corresponds to smooth predictions (constant in these plots) for both the solution and its gradient, leads \ginn~to accurately predict $\sol$ within the training domain $[-1,2]$ (left plot). Similar results are obtained with a \snn with $\ell_2$. This is the result of gradient regularization which can be appreciated by looking at the right plot and noticing how both \ginn~and the \snn~with $\ell_2$ better fit the ground truth gradient respect to \snn. However, while $\ell_2$ induces a reduction in the gradient's magnitude, it does not prevent it from oscillating, see green curve behaviour with respect to the black dashed line on the input space boundaries (around $x<-1$ and $x>2$). Instead, \ginn~reduces the magnitude of the gradient but also penalises its oscillations via the term $\mathcal{L}_\netauxg$.
A similar behaviour is observed when initializing $\netsol$ to have less smooth predictions for $\sol$ but keeping the initialization of $\netauxg$ unchanged, see Fig.~\ref{fig:uwiggly_comparisons} (top row). On the contrary, initializing $\netauxg$ to give less smooth predictions, leads \ginn~to converge to a less smooth solution (Fig.~\ref{fig:uwiggly_comparisons}, bottom row) in line with the interpretation of $\netauxg$ as expressing prior beliefs. See Appendix~\ref{net_init}  for further details.

\textbf{Higher order derivatives} \ginn~can be easily extended to incorporate constraints on higher-order derivatives. For instance, we can constrain the second order derivatives by introducing a third network, \( \netauxgso(\cdot, \Theta_\netauxgso) \) $:\mathbb{R}^d \to \mathbb{R}^{d \times d}$ and considering the augmented loss \(\mathcal{L}(\Theta_{\netsol},\Theta_{\netauxg},\Theta_{\netauxgso}) = \mathcal{L}_{\netsol}(\Theta_{\netsol}) + \mathcal{L}_\netauxg(\Theta_{\netsol},\Theta_{\netauxg}) +\mathcal{L}_\netauxgso(\Theta_{\netsol},\Theta_{\netauxgso})\) where \(\mathcal{L}_\netauxgso\) is given by:
\begin{align}
\mathcal{L}_\netauxgso(\Theta_{\netsol},\Theta_{\netauxgso}) = \frac{1}{\ndataCg} \sum_{\indexC=1}^{\ndataCg}\sum_{i,j=1}^\xdim \left(\netauxgso_{ij}(\vecx_m; \Theta_\netauxgso) - \netsol_{x_ix_j}( \vecx_m; \Theta_U) \right)^2
\label{eq:GINN_eq_loss_2grad}
\end{align}
with $\netauxgso_{ij}$ giving the $(i, j)$-output of the second auxiliary network and $\netsol_{x_ix_j}$ representing the second order partial derivative of $\netsol$ with respect to $x_i$ and $x_j$. An equivalent formulation can be written by using a unique network $\netauxg$~with increased output dimension
but still considering an additional loss component for each gradient order to be constrained.

\textbf{Multi-outputs system}
Finally, \ginn{s} can also be used to approximate multi-output systems in which $\sol(\cdot):\mathbb{R}^\xdim \to \mathbb{R}^\udim$. In this setting, we have $\netsol(\cdot;\Theta_{\netsol}):\mathbb{R}^\xdim \to \mathbb{R}^\udim$ and $\netauxg(\cdot;\Theta_{\netauxg}):\mathbb{R}^\xdim \to \mathbb{R}^{\xdim \times \udim}$ and write $\mathcal{L}_{\netsol}$ and $\mathcal{L}_\netauxg$ as follow:
\begin{align}
\mathcal{L}_{\netsol}(\Theta_{\netsol}) &= \frac{1}{\ndataTg} \sum_{\indexT=1}^{\ndataTg}\sum_{\indexu=1}^{\udim} \left(\netsol^{\indexu}(\vecx_\indexT; \Theta_{\netsol}) - \sol^{\indexu}_\indexT\right)^2,
\label{eq:GINN_eq_loss_MSE_multi} \\
\mathcal{L}_\netauxg(\Theta_{\netauxg}, \Theta_{\netsol}) &= \frac{1}{\ndataCg} \sum_{\indexC=1}^{\ndataCg}\sum_{i=1}^\xdim\sum_{\indexu=1}^{\udim}  (\netauxg^\indexu_{i}(\vecx_\indexC; \Theta_\netauxg)- \netsol^\indexu_{x_i}( \vecx_\indexC; \Theta_{\netsol}) )^2.
\label{eq:GINN_eq_loss_grad_multi}
\end{align}
where \(\netsol^\indexu\) and $\sol^k_n$ are the $k$-th network and ground truth output respectively, \(\netsol^\indexu_{x_i}\) is the derivative of the $\indexu$-th output with respect to the $i$-th input and \(\netauxg^\indexu_{i}\) is the corresponding output of the auxiliary network.

\section{Experiments}
We test \ginns~on a well known synthetic function, that is the Friedman function (\friedman, $\xdim=5$, $\udim=1$), and three physical systems featuring different characteristics in terms of time dependency, input and output dimension and smoothness of the gradients:
\begin{itemize}
    \item the Stokes Flow (\stokes), which can be used to describe the motion of fluids around a sphere, in conditions of low Reynolds number \((Re < 1)\). In this experiment $\udim=1$, there is no dependency of $\sol$ on time and $\xdim=2$.
    \item The Lotka-Volterra system (\lv), an \acro{ode} model of predator-prey dynamics in ecological systems. This system's $\sol$ depends on time, which is the only input ($\xdim=1$), and $\udim=2$.
    \item The Burger's equation (\burgers), a challenging \acro{pde} studied in fluid mechanics to represents shock waves and turbulence. In this case $\sol$ depends on time and a spatial variable ($\xdim=2$) and $\udim=1$. This equation is particularly challenging due to steep partial derivatives within the solution's domain.
\end{itemize}

\textbf{Metrics} Given a test dataset $\mathcal{D}_{\acro{test}} = \{\vecx_\indexTest, \sol_\indexTest\}_{\indexTest=1}^{\ntest}$, \ginn's performance is assessed in terms of root mean squared error of both output ($\rmse_U$\footnote{$\rmse_{\netsol} = \sqrt{\frac{1}{\ntest} \sum_{\indexTest=1}^{\ntest} (\netsol_\indexTest - \sol_\indexTest)^2}$ where $\sol_\indexTest$ and ${\netsol}_\indexTest$ are the ground truth solution and \ginn's predicted output for the $\indexTest$-th datapoint $\vecx_\indexTest$ in $\mathcal{D}_{\acro{test}}$ respectively.} ) and gradient predictions ($\rmse_\partial$\footnote{
$\rmse_\partial = \sqrt{\frac{1}{\ntest} \sum_{\indexTest=1}^{\ntest} (\netsol_{x_i,\indexTest} - \sol_{x_i,\indexTest})^2}$ where $\netsol_{x_i, \indexTest}$ and $\sol_{x_i, \indexTest}$ are \ginn's gradient prediction and the ground truth gradient with respect to the $i$-th dimension $x_i$ of a given test point $\vecx_\indexTest$ respectively.}). In order to compute $\rmse_\partial$, we derive the ground truth gradient analytically via closed-form differentiation of the original function (for \friedman) or through the \acro{ode} (for \lv). When analytical computation of the gradient is not possible, this is obtained via automatic differentiation of a standard \nn that has been trained on a large dataset ($\ndataTg>10^4$) to ensures accurate predictions for both the solution and its gradient (for \stokes~and \burgers).

\textbf{Baselines}  
As \ginn~does not rely on prior physical knowledge, we mainly compare it against \snn{s}. However, for completeness of our experimental results, we also report figures obtained with Sobolev training \cite{sobolev} (denoted by \acro{st} henceforth), \acro{upinn} and \acro{dhpm} despite these models assume a certain level of prior knowledge of the governing equations and its gradient.
%
In particular, \acro{st} assumes the availability of a training dataset that also includes the ground truth values of the gradient for each $\vecx \in \mathcal{D}_{\acro{train}}$. \acro{dhpm} assumes known inputs of the auxiliary network $\netauxp$ (see Eq.\eqref{eq:PINN_eq_f_upinn}). Finally, \acro{upinn} assumes, on top of the inputs of $\netauxp$, partial knowledge of the functional form of $\diffoperator$.
Note that, for both \acro{upinn} and \acro{dhpm}, we do not re-run the code associated to these two methodologies but use the figures reported in the corresponding papers \cite{Raissi_2018, 10.5555/3618408.3619569} directly. 

\textbf{Experimental details} 
The network architectures of both \(\netsol\) and \(\netauxg\) are fixed across all experiments. In particular, we set \(\netsol\) to be a network with three hidden layers including 20 neurons each. For \(\netauxg\) we consider two hidden layers comprising 50 neurons each. Both \(\netsol\) and \(\netauxg\) use sigmoid activation functions across all layers except the final one, which employs a linear activation function. \(\netsol\) and \(\netauxg\) are trained simultaneously for \(10^{4}\) epochs. For $\netsol$, a fixed batch size ($bs_{\netsol}$) of 64 is used for all experiments apart from \lv~where $bs_{\netsol}=5$. For \(\netauxg\), the batch size $bs_\netauxg$ is set according to $bs_\netauxg = bs_{\netsol}\times \ndataCg/\ndataTg$. The \snn~used for comparison uses the same network architecture for \(\netsol\) and is trained for the same number of epochs. We use Adam optimizer \cite{adam2015} with a learning rate determined by an inverse time decay schedule, where the learning rate starts at 0.1 and continuously decreases over 500 epochs based on a decay rate of 0.90. Across all experiment, both \netsol~and \netauxg~are initialized with biases set to zero and weights with \emph{Glorot uniform} distribution (see \cite{Glorot2010_GLOROTunifANDnormal}). This leads to constant network outputs (for both $\netsol$ and $\netauxg$) at initialization which imply smooth prior beliefs.\footnote{All experiments were performed on a local machine with 6 \acro{cpu}{s} (2.10GHz, 16\acro{gb} of \acro{ram}).} 

\subsection{\friedman}
In this section we assess \ginn's performance on the five dimensional Friedman function \cite{friedman1991multivariate}:
\begin{equation}
u(\vecx) = 10 \sin(\pi x_1 x_2) + 20 (x_3 - 0.5)^2 + 10 x_4 + 5 x_5 + \epsilon
\label{eq:friedman_eq}
\end{equation}
where $\epsilon \sim \mathcal{N}(0, \sigma^2)$ with $\sigma^2$ taking different values across experiments. We construct $\dataTg$ and $\dataCg$ by sampling input points (via Latin hypercube sampling) in the 5-dimensional unit hypercube and obtaining the corresponding output values via Eq. \eqref{eq:friedman_eq}. The test dataset includes \(\ntest = 10^{4}\) points uniformly sampled in the 5-dimensional unit hypercube. We first demonstrate our approach on noise-free data ($\sigma^2=0$) and then explore settings with increasing level of $\sigma^2$.

\begin{figure}[t]
\centering
\begin{minipage}{0.48\textwidth}
\centering
\includegraphics[width=\linewidth]{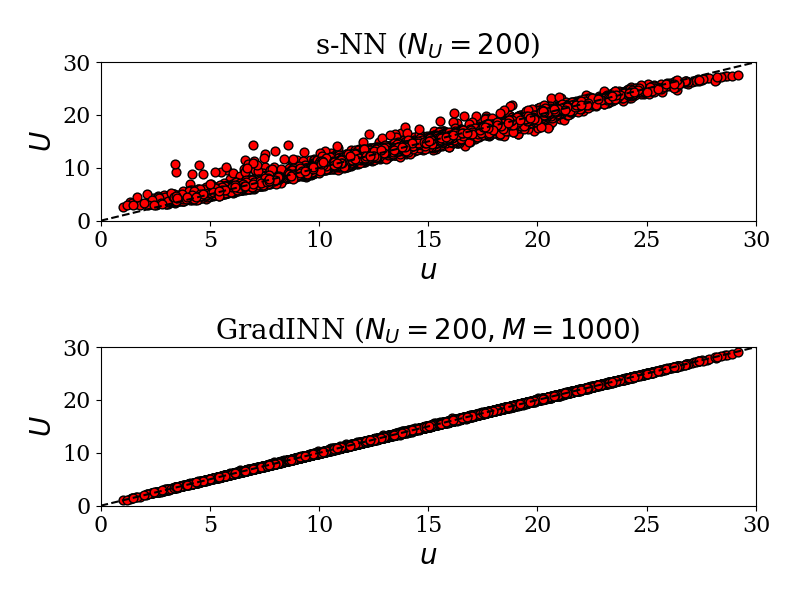}
\captionof{figure}{\friedman. Predicted output $\netsol$ vs $\sol$ for each \(\vecx_n \in \dataTp_{\acro{test}}\) \textit{Top}: \snn~with $\rmse_U = 0.51$. \textit{Bottom}: \ginn~with $\rmse_U = 0.04$.}
\label{fig:Friedman_U_regression}
\end{minipage}%
\hfill
\begin{minipage}{0.48\textwidth}
\centering
\captionof{table}{\friedman. $\rmse_U$ for different values of $\ndataTg$ and $\ndataCg$ when $\sigma^2=0$. For each value of $\ndataTg$ the lowest \rmse is bolded.}
\resizebox{\columnwidth}{!}{
\begin{tabular}{lccccc}
\toprule
& & \multicolumn{4}{c}{$\ndataTg$} \\
\cmidrule{3-6}
& & 50 & 100 & 200 & 500 \\
\midrule
\multirow{3}{*}{\snn} & - & 2.24 & 1.55 & 0.51 & \textbf{0.02} \\
                      & $\ell_1$ & 1.48 & 0.61 & 0.13 & 0.05 \\
                      & $\ell_2$ & 1.52 & 0.48 & 0.13 & 0.06 \\
\cmidrule{1-6}
\multirow{3}{*}{\ginn} & $M=500$ & 0.77 & 0.24 & 0.07 & \textbf{0.02} \\
                       & $M=10^{3}$ & 0.75 & 0.22 & \textbf{0.04} & \textbf{0.02} \\
                       & $M=10^{4}$ & \textbf{0.66} & \textbf{0.19} & 0.05 & \textbf{0.02} \\
\cmidrule{1-6}
\multirow{1}{*}{\acro{st}} & & 1.50 & 0.50 & 0.17 & \textbf{0.02} \\
\bottomrule
\label{RMSE_predictions_noise_free}
\end{tabular}}
\end{minipage}
\label{}
\end{figure}


\begin{table}[t]
\caption{\friedman. $\rmse_\partial$ for different values of $\ndataTg$ and across input dimensions when $\sigma^2=0$.}\label{RMSE_NN_grad}
\resizebox{\columnwidth}{!}{
\begin{tabular}{lccccccccccccccc}
\toprule
&  \multicolumn{5}{c}{\snn} & \multicolumn{5}{c}{\ginn~($\ndataCg=10^3$)} & \multicolumn{5}{c}{\acro{st}} \\
\cmidrule(lr){2-6} 
\cmidrule(lr){7-11} 
\cmidrule(lr){12-16} 
\(\ndataTg\) & \(\netsol_{x_1}\) & \(\netsol_{x_2}\) & \(\netsol_{x_3}\) &  \(\netsol_{x_4}\) &  \(\netsol_{x_5}\) & \(\netsol_{x_1}\) &\(\netsol_{x_2}\) & \(\netsol_{x_3}\) & \(\netsol_{x_4}\) & \(\netsol_{x_5}\) & \(\netsol_{x_1}\) & \(\netsol_{x_2}\) & \(\netsol_{x_3}\) & \(\netsol_{x_4}\) & \(\netsol_{x_5}\) \\
\cmidrule(lr){1-1} 
\cmidrule(lr){2-6}
\cmidrule(lr){7-11}
\cmidrule(lr){12-16}
50    & 10.30 & 9.50 & 11.10 & 5.88 & 3.99  & 3.68 & 4.44 & 3.84 & 0.71 & 0.69 & 7.60 & 8.20 & 7.90 & 3.40 & 3.30\\
100   & 7.30  & 7.80 & 8.20 & 5.30 & 4.33  & 1.83 & 1.90 & 0.45 & 0.25 & 0.10 & 3.50 & 3.50 & 2.60 & 1.40 & 1.30\\
200   & 3.97 & 3.66 & 3.35 & 2.07 & 1.74  & 0.44 & 0.48 & 0.42 & 0.09 & 0.05 & 2.10 & 2.30 & 1.01 & 0.72 & 0.39\\
500   & 0.29 & 0.25 & 0.30 & 0.1 & 0.05  & 0.22 & 0.21 & 0.15 & 0.03 & 0.02 & 0.52 & 0.30 & 0.20 & 0.10 & 0.05\\
\bottomrule
\end{tabular}
}
\end{table}

We compare \ginn~with varying number $\ndataCg$ of collocation points against \snn, \snn~with $\ell_1$ and $\ell_2$ regularization\footnote{We choose $\ell_1$ and $\ell_2$ values to be those giving the lower $\rmse_{\netsol}$ values over $\dataTp_{\acro{test}}$. The figures shown are thus a lower bound on the \rmse achieved with $\ell_1$ or $\ell_2$ regularization. Note that considering  $\ell_1$ or $\ell_2$ for $\ndataTg=500$ leads to increased \rmse.} and \acro{st}, despite the different training dataset considered by the latter. 
Table~\ref{RMSE_predictions_noise_free} and Table~\ref{RMSE_NN_grad} show the \rmse performances achieved in the noiseless data case ($\sigma^2=0$) for the output and gradient predictions respectively and across different values of $\ndataTg$ ($\rmse_{\partial}$ for \snn~with $\ell_1$ and $\ell_2$ regularization are reported in Table \ref{RMSE_NN_grad_LT_l1_l2} in the appendix). Notice how, for a fixed value of $\ndataTg$, \ginn~significantly outperforms \snn~in terms of output predictions both with and without regularization (see Fig.~\ref{fig:Friedman_U_regression} for a visualization of the difference between the ground truth and the predicted output values over the test dataset). As expected, the use of an increasing $\ndataCg$ further decreases the $\rmse_U$ in settings where $\ndataTg$ is low. In addition, note how \ginn~outperforms \acro{st} without having access to the true gradient values for the training points.
More importantly, \ginn~displays high performance in terms of gradient predictions across all input dimensions and all settings of $\ndataTg$ (Table 2). This can be further appreciated by looking at Fig.~\ref{fig:Friedman_U_grad_tot} in the appendix which shows 
the gradient prediction $\nabla_{\vecx} \netsol$ for both \snn~and \ginn. 
As expected, when \(\dataTg\) is sufficient to characterize the system ($\ndataTg$ is high) the performance of \snn~matches the one achieved by \ginn~both in terms of $\rmse_U$ and $\rmse_{\partial}$. When instead $\ndataCg=0$, \ginn~simplifies (\(\mathcal{L}_\netauxg=0\)) thus approaching \snn.

We repeat the same analysis with fixed values of \(\ndataTg=200\) and \(\ndataCg=1000\) but an increasing level of noise in  $\mathcal{D}_{\acro{train}}$. 
Denote by $\sigma_u$ the standard deviation of the output values in the $\mathcal{D}_{\acro{train}}$ and let $c$ be a constant taking values in $[0.01, 0.03, 0.05]$. We set $\epsilon \sim \mathcal{N}(0, (c\sigma_u)^2)$ so as to make the noise proportional to the variability of the output values thus simulating real-world data imperfections and testing the \ginn's robustness. 
Although all models exhibit a decreasing performance for increasing $c$, \ginn~demonstrates higher accuracy across all noise levels thus being more robust to e.g. measurement errors (Table~\ref{RMSE_noise_tab} in Appendix \ref{Appendix_Friedman}). 

\subsection{\stokes}
\begin{figure}[t]
\centering
\begin{minipage}{0.55\textwidth}
\centering
\includegraphics[width=0.9\linewidth]{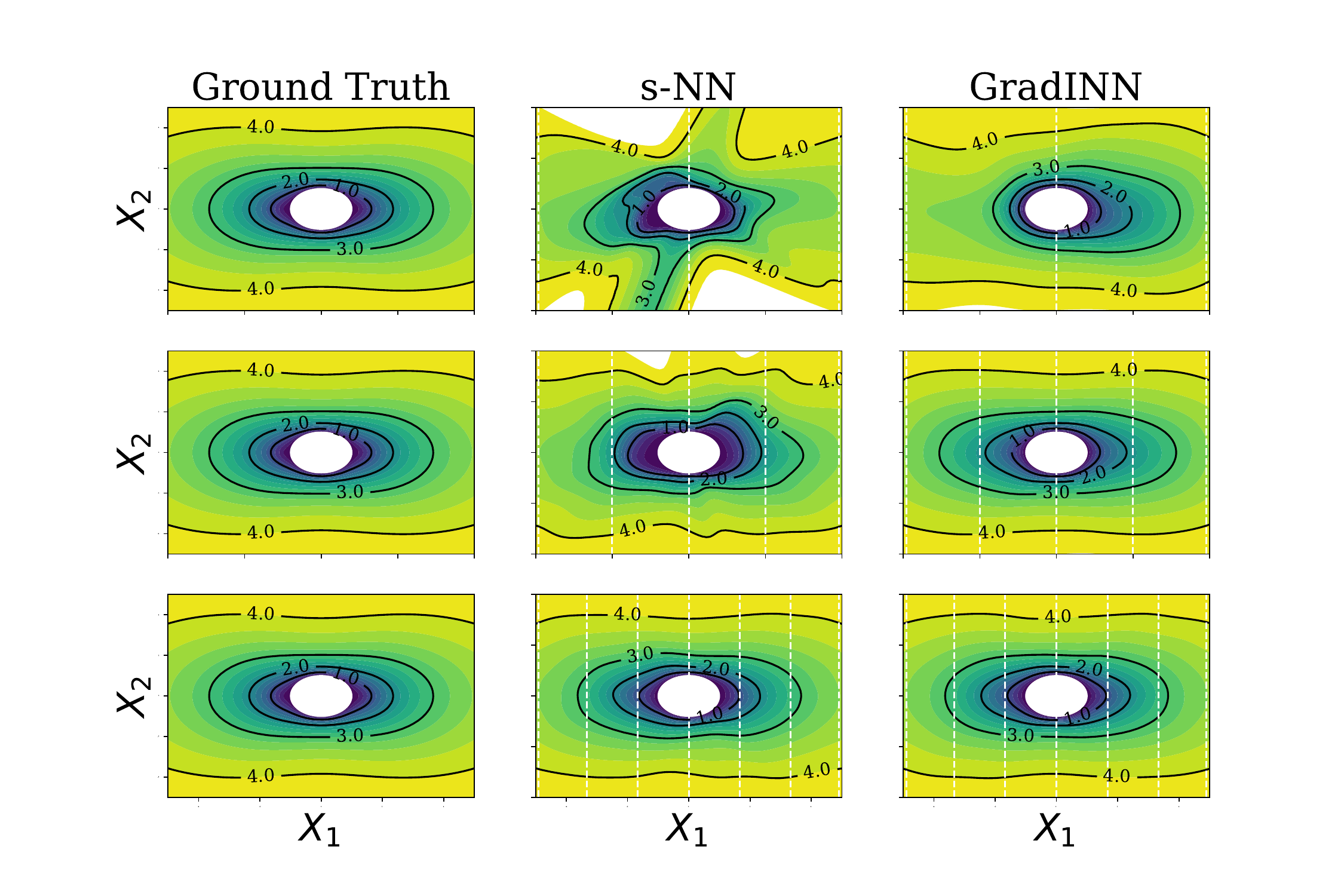}
\vspace{-0.3cm}
\captionof{figure}{\stokes. Predicted solution and ground truth values (first column) for different $\ndataTg$ values (white dotted lines). \textit{Top:} $\ndataTg = 350$. \textit{Middle:} $\ndataTg = 550$. \textit{Bottom:} $\ndataTg = 750$.}
\label{stokes_fig}
\end{minipage}%
\hfill
\begin{minipage}{0.42\textwidth}
\centering
\captionof{table}{\stokes. $\rmse_{\netsol}$ and  $\rmse_{\partial}$ for different $\ndataTg$. For each $\ndataTg$, the lowest \rmse values are bolded.}
\label{stokes_tab}
\begin{small}
\begin{tabular}{llccc}
\toprule
& & \multicolumn{3}{c}{\(\ndataTg\)} \\
\cmidrule{3-5}
 &  & 350 & 550 & 750 \\
\midrule
\multirow{2}{*}{\(\netsol\)} & \snn & 0.37 & 0.20 & 0.04 \\
                             & \ginn & \textbf{0.24} & \textbf{0.05} & \textbf{0.03} \\
                             \midrule
\multirow{2}{*}{\(\netsol_{x_1}\)} & \snn & 0.51 & 0.31 & 0.09 \\
                                  & \ginn & \textbf{0.27} & \textbf{0.09} & \textbf{0.05} \\
                                  \midrule
\multirow{2}{*}{\(\netsol_{x_2}\)} & \snn & 0.44 & 0.32 & 0.08 \\
                                  & \ginn & \textbf{0.17} & \textbf{0.07} & \textbf{0.07} \\
\bottomrule
\end{tabular}
\end{small}
\end{minipage}
\label{Sphere_solution}
\end{figure}

Next, we test our proposed methodology on the Stokes flow \cite{Stokes}, also known as creeping flow or viscous flow. This regime describes fluid motion at very low Reynolds numbers, where inertial forces are negligible compared to viscous forces. In the case of a sphere of radius \( R \) (white circle in Fig.~\ref{stokes_fig}) moving through a fluid with relative far-field velocity \( \hat{u}_{\infty} \), the Stokes flow permits a closed-form solution \( u(\vecx) \in \mathbb{R}^3\) at any point \( \vecx =[x_1, x_2, x_3]^T\). 
Denote by $\vecx \otimes \vecx$ the tensor product of $\vecx$ with itself, by $\|\vecx\|$ the Euclidean norm of $\vecx$ and by $\mathbf{I}$ the identity matrix, the Stoke flow is defined as: 
\begin{align*}
u(\vecx) = \bigg( &\frac{3 R^3}{4} \frac{\vecx \otimes \vecx}{\|\vecx\|^5} - \frac{R^3}{4} \frac{\mathbf{I}}{\|\vecx\|^3}- \frac{3 R}{4} \frac{\vecx \otimes \vecx}{\|\vecx\|^3} - \frac{3 R}{4} \frac{\mathbf{I}}{\|\vecx\|} + \mathbf{I} \bigg) \times \hat{u}_{\infty}.
\label{Stokes flow}
\end{align*} \begin{wrapfigure}[16]{r}{0.4\textwidth}
\vspace{-0.45cm}
\centering
\includegraphics[width=0.35\columnwidth]{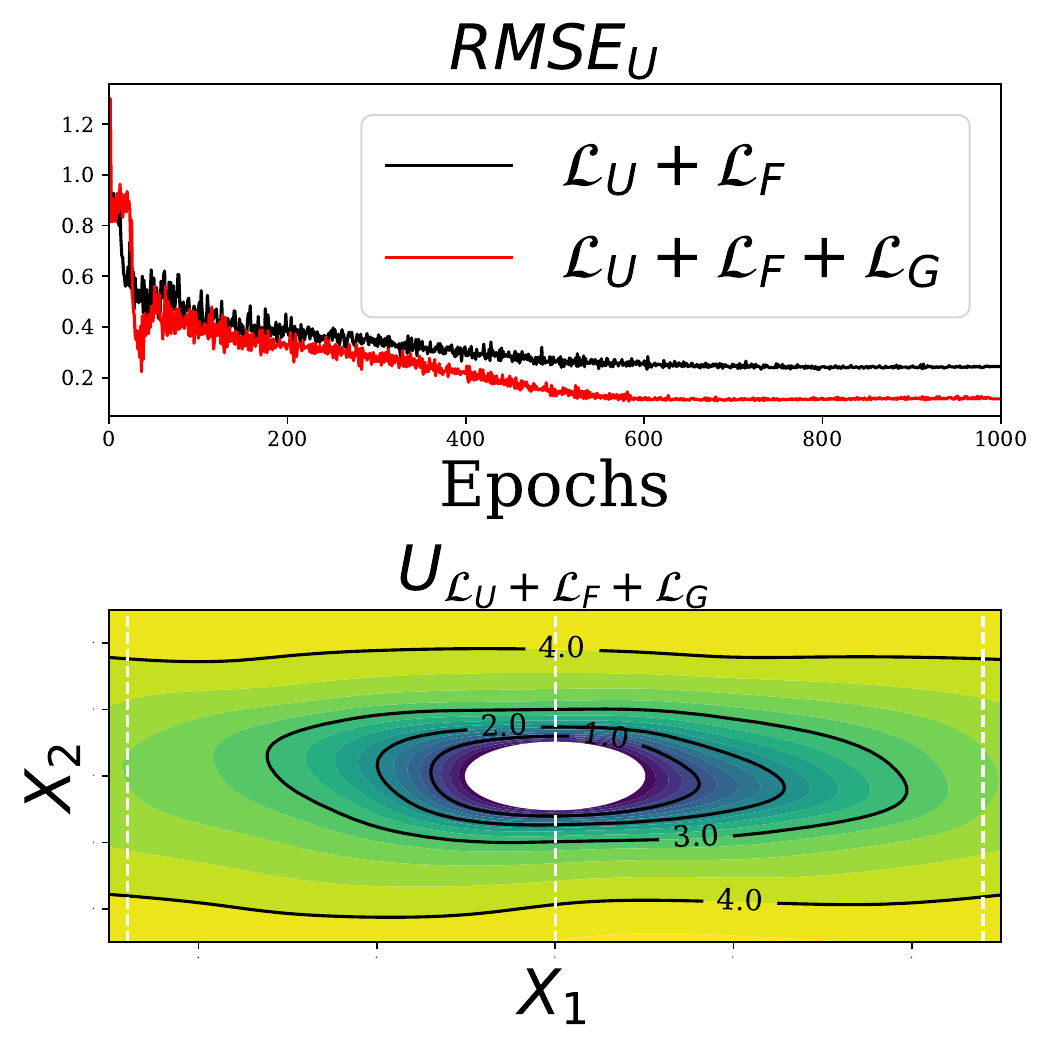}
\caption{\stokes. \textit{Top}: $\rmse_U$ over training epochs when constraining $\nabla_{\vecx}\netsol$ (black line) and both the $\nabla_{\vecx}\netsol$ and $\nabla^2_{\vecx}\netsol$ (red line). \textit{Bottom}: predicted solution when constraining $\nabla_{\vecx}\netsol$ and $\nabla^2_{\vecx}\netsol$ ($\ndataTg=350$).}
\label{Sphere_solution_second_order}
\end{wrapfigure}

We reduce the input dimension for $\vecx$ to $d=2$ by fixing $x_3=0$ and considering input values, for both $x_1$ and $x_2$, in $[-5, 5]$. Similarly, we reduce the output dimension to one by taking the Euclidean norm \( \|\mathbf{u}\| \)\footnote{\( \|\mathbf{u}\| = \sqrt{u_1^2 + u_2^2 + \ldots + u_n^2} \) for \( \mathbf{u} = (u_1, u_2, \ldots, u_n) \).} of each output vector and training both \ginn~and \snn~on those. We fix \(\hat{u}_{\infty}=[5, 0, 0]^T\). We generate $\dataTg$ by considering a regular grid of input values of size $\ndataTg \in [300, 500, 700]$ (see dotted white lines in Fig.~\ref{stokes_fig}). As the homogeneous Dirichlet boundary conditions are assumed to be known, we add 50 points along the sphere's perimeter section with \(\|\mathbf{u}\| = 0\) to this training dataset.  $\dataCg$ is constructed by taking \( \ndataCg=10^4 \) points in the $[-5, 5]^2$ domain. Similarly, \(\dataTp_{\acro{test}}\) includes a regular grid of points $N_{\acro{test}} = 4 \times 10^4 $ in the same domain.
\ginn~outperforms \snn~across all $\ndataTg$ values, both in terms of output and gradients predictions (Table~\ref{stokes_tab}). In particular, \snn~exhibits a behaviour similar to the ground truth  solution only when trained with \(\ndataTg = 750\) (Fig.~\ref{stokes_fig}, bottom row). On the contrary \ginn~successfully reconstructs the ground truth solution starting from \(\ndataTg = 550\) (see Fig.~\ref{stokes_fig}, middle and bottom rows).

We further test \ginn~by applying the loss given in Eq. \eqref{eq:GINN_eq_loss_2grad} to constrain both $\nabla_{\vecx}\netsol$ and $\nabla^2_{\vecx}\netsol$. We fix $\ndataTg = 350$ and keep the same $\dataCg$. 
Fig.~\ref{Sphere_solution_second_order} (top) shows how incorporating the additional $\mathcal{L}_{\netauxgso}$ decreases $\rmse_{\netsol}$ (from 0.24 to 0.11) leading to a predicted solution (Fig.~\ref{Sphere_solution_second_order}, bottom) closer to the ground truth (see first col of Fig.~\ref{stokes_fig}). This further highlights the benefit and flexibility of the proposed methodology. The same procedure was followed for $\ndataTg = 550$ and $\ndataTg = 750$ but given the already high accuracy of \ginn~reached with these sizes of $\dataTp_{\acro{train}}$ using only \(\mathcal{L}_{\netsol}+\mathcal{L}_\netauxg\), no significant further improvement was observed in terms of both $\rmse_U$ and $\rmse_\partial$ when constraining the second order gradient.
Note that, although the $\rmse_U$ stabilises within 1000 epochs in both configurations (Fig.~\ref{Sphere_solution_second_order}, top), the computational effort significantly increases when constraining the second-order gradient leading to higher training time.


\subsection{\lv}
\label{sec:Lotka-Volterra}
\begin{figure}[t]
\centering
\begin{minipage}{0.65\textwidth}
\centering
\includegraphics[width=\linewidth]{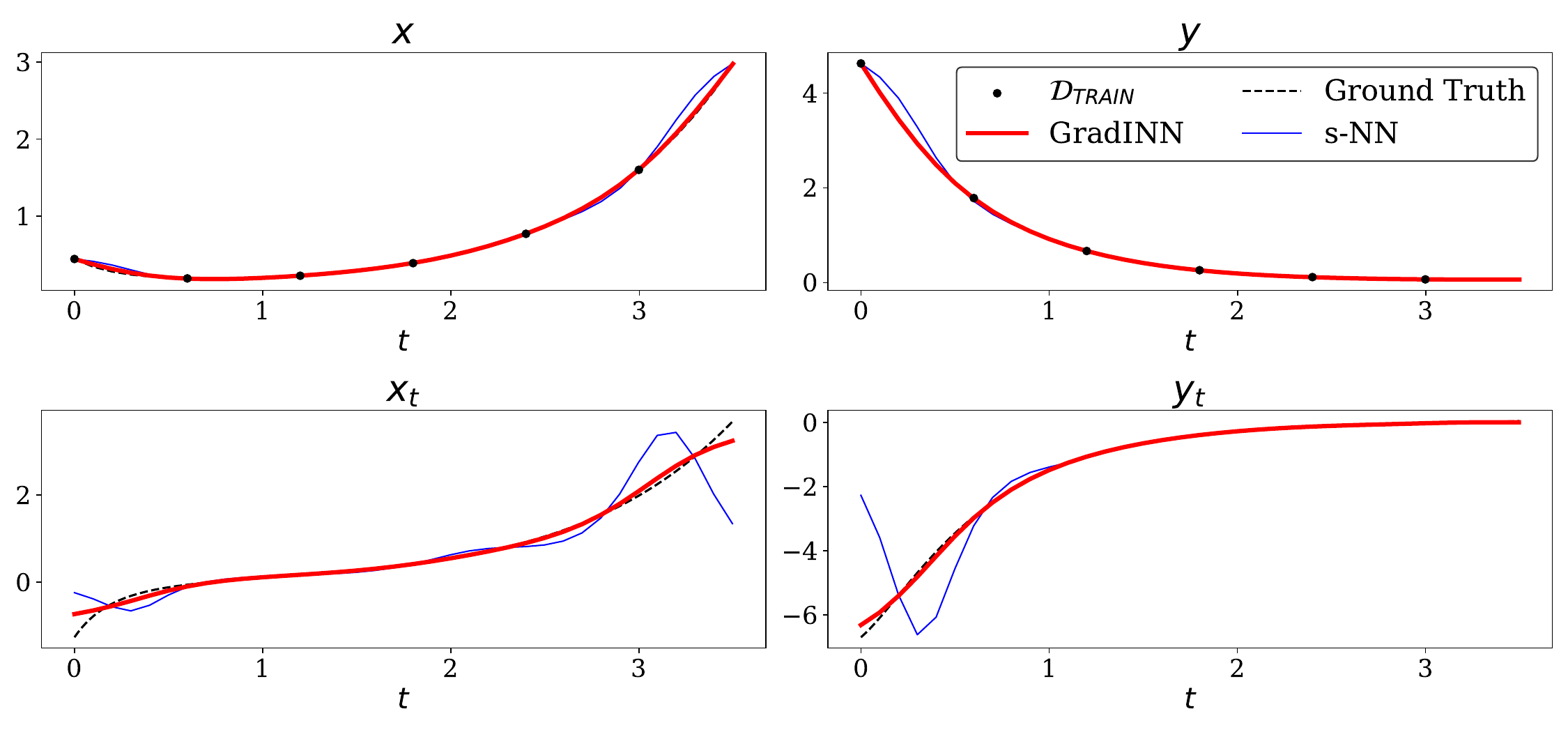}  
\captionof{figure}{\lv. \textit{Top}: Predicted solution. \textit{Bottom}: Predicted gradient.}
\label{lv_fig}
\end{minipage}%
\hfill
\begin{minipage}{0.3\textwidth}
\centering
\captionof{table}{\lv. $\rmse_\partial$ for the predator’s uptake functions i.e. \(- \beta xy\) and \(\delta xy\), the lowest \rmse values are bolded.}
\label{lv_tab}
\begin{small}
\begin{tabular}{lcc}
\toprule
Method & & \rmse\\
\midrule
\upinn & - & \textbf{0.03}\\
\multirow{2}{*}{\snn} & \(- \beta xy\) & 0.11\\
& \( \delta xy\) & 0.16\\
\multirow{2}{*}{\ginn} & \(- \beta xy\) & \textbf{0.03}\\
& \( \delta xy\) & \textbf{0.025}\\
\bottomrule
\end{tabular}
\end{small}
\end{minipage}
\label{LV}
\end{figure}

Moving to time-dependent systems, we test \ginn~on the Lotka-Volterra (\lv) system \cite{LV}, described by the following \textsc{ode}s:
\begin{equation}
    x_t = \alpha x - \beta xy \quad \quad
    y_t = \delta xy - \gamma y,
    \label{eq:lv}
\end{equation}
where \( x \) and \( y \) represent the populations of two species, typically prey and predators respectively. The parameters \( \alpha \) and \( \beta \) characterize the prey dynamics while \( \gamma \) and \( \delta \) characterize the predator dynamics. As in \citet{10.5555/3618408.3619569}, we numerically solve the system for \([\alpha, \beta, \gamma, \delta] =[1.3, 0.9, 0.8, 1.4] \) with initial condition \([x_0, y_0] = [0.44249296, 4.6280594]\) and exploit the solution to generate $\dataTg$ with $\ndataTg=5$ and uniformly distributed input values in \(t \in [0, 3]\). For \ginn~we set \(\ndataCg=1000\). \ginn~gives more accurate predictions compared to \snn, both in terms of the predicted output (Fig. \ref{lv_fig}, top) and gradients (Fig. \ref{lv_fig}, bottom).
%
We compare the gradient predictions obtained with \ginn~against the results reported in \cite{10.5555/3618408.3619569}, where \upinn~is used to discover only part of the system in Eq. \eqref{eq:lv} , namely the predator’s uptake functions, i.e., \(- \beta xy\) and \(\delta xy\).  Using the same $\dataTg$ and number of collocation points (\(\ndataCg = 1000\)), \upinn~achieves an average $\rmse_\partial$ across the unknown parts of the gradients of $0.03$\footnote{It is not clear how the \rmse of the predictions for \(- \beta xy\) and \(\delta xy\) is computed in \cite{10.5555/3618408.3619569}. }. Instead, \ginn's average $\rmse_\partial$ across output dimensions is $0.08$. The difference in these \rmse values can be attributed to the absence of prior knowledge considered by \ginn. Indeed, this value of $\rmse_\partial$ also includes the predictions for the parts of  Eq. \eqref{eq:lv} that are considered known in \upinn. To facilitate the comparison of \ginn~and \upinn~and have a more comparable metric, we evaluate the \(\rmse_\partial\) only for the predator’s uptake functions (Table \ref{lv_tab}). In this case, \ginn~prediction accuracy over \(- \beta xy\) and \(\delta xy\) is comparable to the one achieved by \upinn. 



\subsection{\burgers}
\label{burger}
Finally, we test \ginn's capability to solve \textsc{pde}s by considering the Burgers' equation \cite{BASDEVANT198623}.
For a given field $\sol(t, x)$ and kinematic viscosity \(\nu\) this is defined as $\sol_t + \sol \sol_{x} = \nu \sol_{xx}$. We consider the initial conditions $u(0, x) = -sin(\pi x/8)$ with $x \in [-8,8]$, $t \in [0,10]$, \(\nu =0.1\) and known Dirichlet boundary conditions. We take the solutions of the \acro{pde} from \cite{Raissi_2018}. This gives a set of $201 \times 256$ points from which we select $\ndataTg=1000$ for $\dataTg$ in $t \in [0, 6.7]$ and keep the rest for $\dataTp_{\acro{test}}$. Note that we select $\mathcal{D}_{\acro{train}}$ to cover approximately two-thirds of the complete input domain thus testing the model capability to generalize outside of the training domain. 
To construct $\mathcal{C}$, we uniformly sample $\ndataCg = 22000$ points in the full inputs domain. 

\begin{figure}[t]
\centering
\begin{minipage}{0.65\textwidth}
\centering
\includegraphics[width=\linewidth]{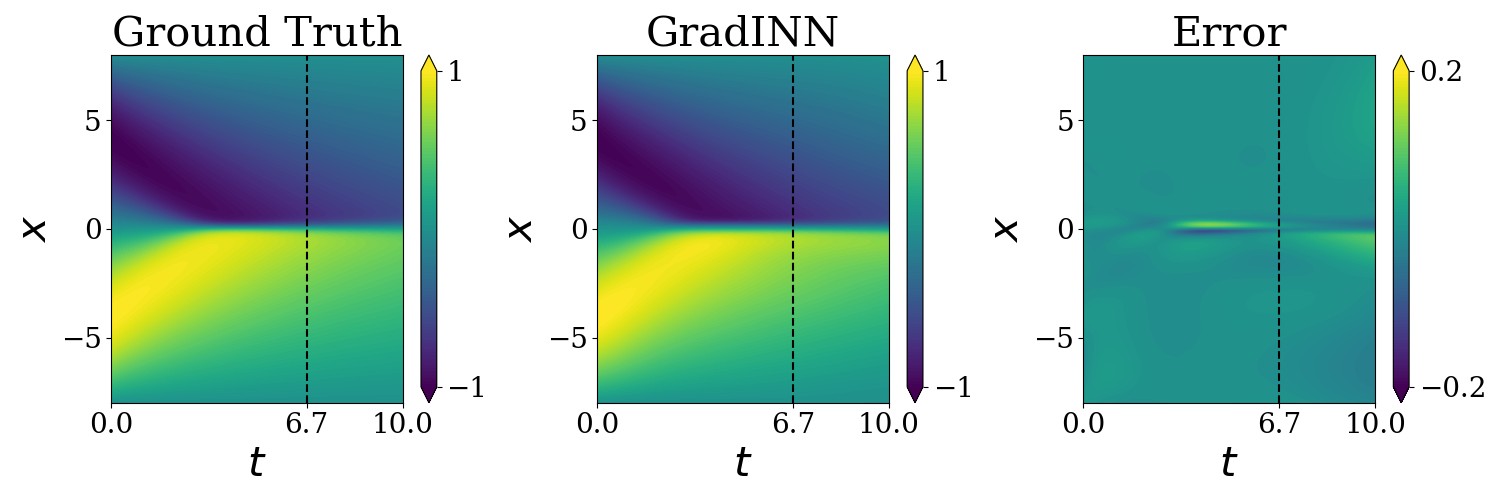}
\caption{\burgers. \textit{Left}: Ground truth solution.  \textit{Middle:} \ginn's predicted solution. \textit{Right:} Difference between \ginn~solution and ground truth values.
Using $\dataTp_\acro{train}$ in $t \in [0, 6.7]$ (left of black line), \ginn~gives \(\rmse_U = 0.02\), \(\rmse_{U_t} = 0.04\) and \(\rmse_{U_x} = 0.02\).}
\label{burgers_fig}
\end{minipage}%
\hfill
\begin{minipage}{0.3\textwidth}
\centering
\vskip -0.1in
\captionof{table}{\burgers. $\mathcal{L}^{\text{r}}_2$ for \ginn, \snn, and \acro{dhpm} with ($\acro{dhpm}^+$) and without ($\acro{dhpm}^-$) prior knowledge on $\diffoperator$'s inputs. The lowest \rmse values is bolded.}
\label{burgers_tab}
\begin{small}
\begin{tabular}{lc}
\toprule
Method & $\mathcal{L}^{\text{r}}_2$ \\
\midrule
\snn & $11\times 10^{-2}$ \\
\ginn & $2.7\times 10^{-2}$ \\
$\acro{dhpm}^+$ & $\textbf{0.48}\times 10^{-2}$ \\
$\acro{dhpm}^-$ & 1.46 \\
\bottomrule
\end{tabular}
\end{small}
\label{Burgers}
\vskip -0.25in
\end{minipage}
\end{figure}

\ginn's predicted solution, along with the ground truth values and the difference between these two 
are given in Fig.~\ref{burgers_fig}. 
This comparison reveals a slight oversmoothing of \ginn's predicted solution near the steep gradient at $x=0$, resulting in a localized increase in error (Fig.~\ref{burgers_fig}, right).\footnote{To further analyze this phenomenon, we run an additional experiment with $\nu = 0.01/\pi$, a configuration that leads to even steeper gradient, and observed a more pronounced oversmoothing, see Appendix \ref{Burger_add_exp}.} 
To directly compare our results against the \acro{dhpm} figures given in \cite{Raissi_2018}, we compute an alternative metric that is the relative \(\mathcal{L}_2\) ($\mathcal{L}^{\text{r}}_2$) error.\footnote{$\mathcal{L}^{\text{r}}_2 = \sqrt{\frac{\sum_{\indexTest=1}^{\ntest} (\sol_\indexTest - \netsol_\indexTest)^2}{\sum_{\indexTest=1}^{\ntest} u_\indexTest^2}}$ where \(u_\indexTest\) and \(\netsol_\indexTest\) give the ground truth and predicted value for $\vecx_h$ respectively. 
} As in the previous experiments, \ginn~outperforms \snn~(Table~\ref{Burgers}) while, as expected, the \acro{dhpm} approach using prior knowledge ($\acro{dhpm}^+$) achieves a higher accuracy. However, it is important to note that, as reported in \cite{Raissi_2018}, this performance depends on the knowledge of the input variables for \( \diffoperator \) (see Eq. \eqref{eq:PINN_eq_f_upinn}). When the prior knowledge of which variables influence the system's dynamics is removed ($\acro{dhpm}^-$), \(\mathcal{L}^{\text{r}}_2 \) increases up to \(1.46\) thus performing significantly worse than \ginn.

\section{Conclusions and discussion}
We introduced \ginn, a general methodology for training \nn~ that 
\textit{(i)} allows regularizing gradient of different orders using prior beliefs, \textit{(ii)} does not require prior knowledge of the system and \textit{(iii)} can be used across setting where input and output have different dimensions. Our extensive experimental comparison showed how \ginn~can be used to accurately predict the behaviour of various physical systems in sparse data conditions. In particular, \ginn~outperforms \snn~and performs similarly to \pinn-like approaches by only using data and prior beliefs. 


\textbf{Limitations} In this work we focused on physical systems and consider smooth networks' initializations.
However, this might not be the optimal initialization for problems that exhibit local discontinuities or regions with very steep gradients. For example, as shown for \burgers (Fig.~\ref{burgers_fig} and \ref{fig:Burgers_lowNU}), a smooth initialization can lead oversmooth predictions in areas where the solution is very steep. This highlights the need to define more complex priors to handle such conditions, as well as an effective way to embed such priors in the network $\netauxg$ (e.g., dedicated pretraining of the auxiliary network $\netauxg$).
Additionally, \ginn~incurs a higher computational cost compared to \snn. Indeed, in addition to the standard forward and backward propagation required to update the weights of the networks, \ginn~needs to differentiate the network to evaluate the gradient $\nabla_{\vecx}\netsol$ which translates into a higher 
computational cost.

\textbf{Future Work}
This work opens up several promising avenues for future research. First, different variations of the proposed loss \( \mathcal{L}(\Theta_U, \Theta_F) \) could be considered. For instance, one could use a weighted sum of $\mathcal{L}_U$ and $\mathcal{L}_F$ with hyperparameters multiplying the two losses thus prioritizing training data vs prior beliefs (or viceversa). The loss could also be futher modified to incorporate a consistency loss as discussed in Appendix \ref{add_loss}. Finally, more work is needed to understand the impact of non-smooth $\netauxg$ initialization, as well as more complex prior beliefs such as highly oscillatory behaviors or discontinuities. This will further enhance the flexibility and applicability of \ginn~in modeling a broader range of complex systems.

\appendix

\section{Different network initializations}
\label{net_init}

\begin{figure}[ht]
    \centering
    \begin{minipage}{0.95\textwidth}
        \centering
        \includegraphics[width=0.95\linewidth]{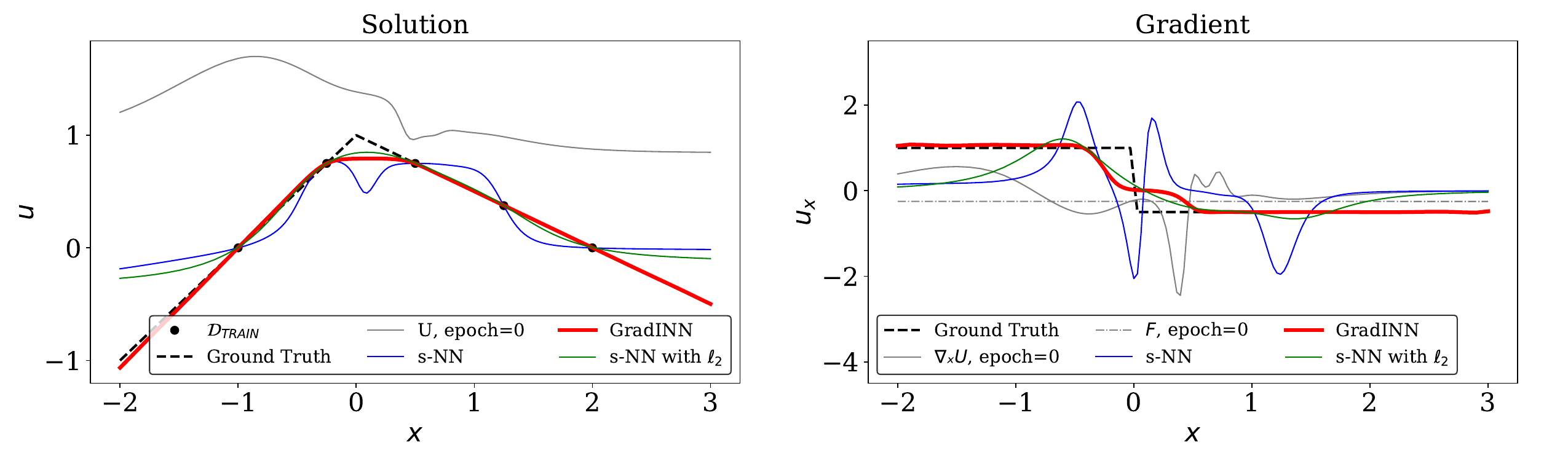}
    \end{minipage}\hfill
    \begin{minipage}{0.95\textwidth}
        \centering
        \includegraphics[width=0.95\linewidth]{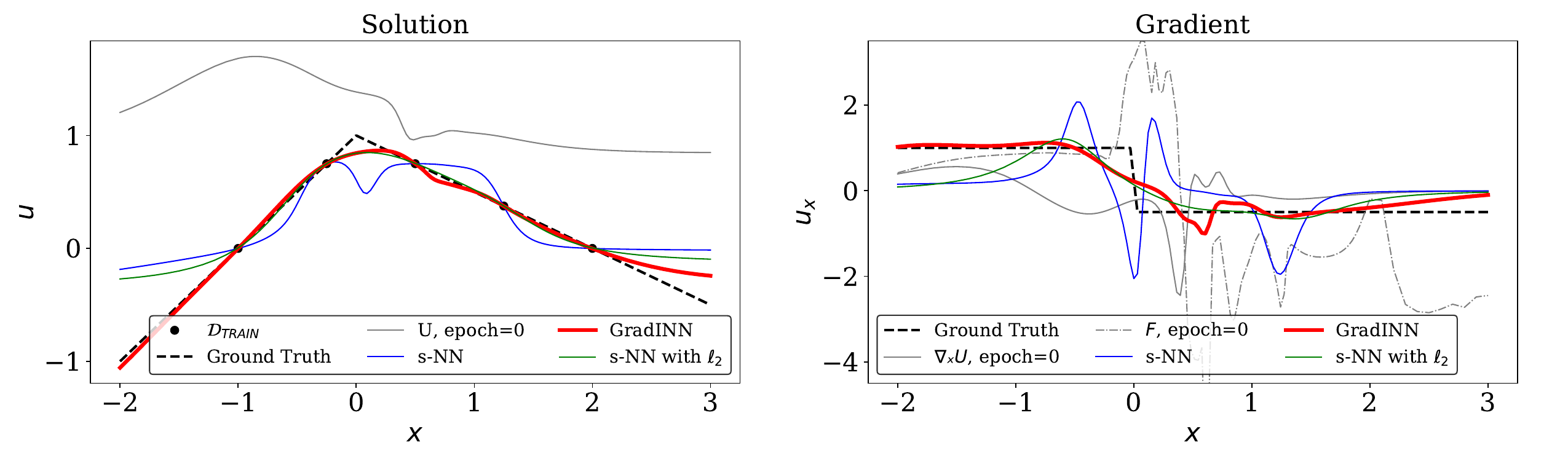} 

    \end{minipage}
    \caption{Ground truth (black dashed lines) and predicted solution (left) and gradient (right) at initialization (gray lines) and after training. Predictions for \ginn~are obtained with $\ndataCg=100$ collocation points uniformly distributed $[-2,3]$. \textit{Top row}: Non-smooth initialization of $\netsol$ and smooth initialization of $\netauxg$.  \textit{Bottom row}: Non-smooth initialization of both $\netsol$ and $\netauxg$.}
    \label{fig:uwiggly_comparisons}
\end{figure}

We repeat the experiment shown in Fig.~\ref{LinSystem_solution} using a different initialization for both \netsol~and \netauxg~to clarify the impact of a non-smooth initializations. We first consider a non-smooth initialization of $\netsol$ but a smooth intialization of $\netauxg$ (top row of Fig.~\ref{fig:uwiggly_comparisons}). In this setting, \snn~converges to a very wiggly solution (blue curve) that improves within the interval $[-1,2]$ when considering the $\ell_2$ norm. When \ginn~is used, similarly to Fig.~\ref{LinSystem_solution}, the smoothness of $\netauxg$ allows recovering a smoother solution with fewer oscillations (red line). When instead the $\netauxg$ initialization is less smooth (dotted grey line in the bottom row of Fig.~\ref{fig:uwiggly_comparisons}), \ginn~recover a less smooth prediction. This is in line with our interpretation of the auxiliary network as expressing prior beliefs.

\section{Additional loss term}
\label{add_loss}
Considering the formulation given in Section \ref{sec:method} for the higher order derivatives, we can embed the second order derivatives with further information adding a consistency loss among the networks $\netauxg$ and $\netauxgso$. In particular, given that $\netauxg$ represents the prior belief on the behaviour of $\nabla_{\vecx}\sol$ and $\netauxgso$ represent the prior belief on the behaviour of $\nabla^2_{\vecx}\sol$,  we can consider the following additional loss term that allows minimizing the residual between $\netauxgso$ and the first-order derivative of $\netauxg$:
\begin{align}
\mathcal{L}_{\netauxgso\netauxg}(\Theta_{\netauxgso}, \Theta_{\netauxg}) = \frac{1}{\ndataCg} \sum_{\indexC=1}^{\ndataCg}\sum_{i,j=1}^\xdim \left(\netauxgso_{ij}(\vecx_m; \Theta_\netauxgso) - \netauxg_{x_j}^i( \vecx_m; \Theta_{\netauxg}) \right)^2\,.
\label{eq:GINN_eq_loss_2grad_consist}
\end{align}
The investigation of the effect of this additional loss term will be part of future work.

\section{Additional results for \friedman}
\label{Appendix_Friedman}

\begin{figure*}[h]
\begin{minipage}{.5\textwidth}
    \centering
    \begin{tikzpicture}
        \node[anchor=south west,inner sep=0] (image) at (0,0) {\includegraphics[width=\linewidth]{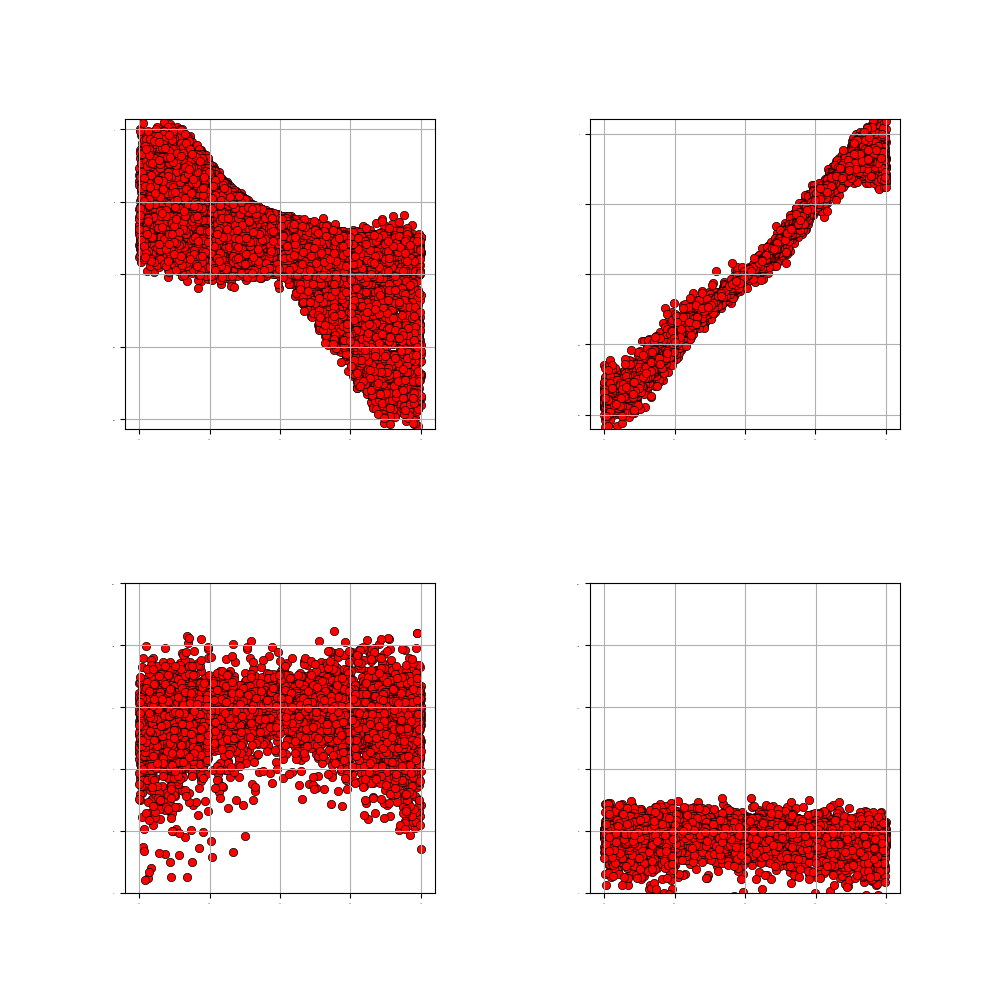}};
        \begin{scope}[x={(image.south east)},y={(image.north west)}]
            \node[font=\large,anchor=south east] at (0.57,0.9) {\snn};
            \node[font=\small,anchor=south east ,rotate=90] at (0.08,0.765) {\(\netsol_{x_1}\)}; 
            \node[font=\small,anchor=south east ,rotate=90] at (0.55,0.765) {\(\netsol_{x_3}\)};
            \node[font=\small,anchor=south east ,rotate=90] at (0.08,0.31) {\(\netsol_{x_4}\)};
            \node[font=\small,anchor=south east ,rotate=90] at (0.55,0.31) {\(\netsol_{x_5}\)};
            \node[font=\small,anchor=south east] at (0.315,0.475) {\(x_1\)};
            \node[font=\small,anchor=south east] at (0.78,0.475) {\(x_3\)};
            \node[font=\small,anchor=south east] at (0.315,0.02) {\(x_4\)}; 
            \node[font=\small,anchor=south east] at (0.78,0.02) {\(x_5\)};
            \node[font=\scriptsize,anchor=south east] at (0.162,0.52) {$0$}; 
            \node[font=\scriptsize,anchor=south east] at (0.315,0.52) {$0.5$}; 
            \node[font=\scriptsize,anchor=south east] at (0.443,0.52) {$1$};
            \node[font=\scriptsize,anchor=south east] at (0.162,0.06) {$0$}; 
            \node[font=\scriptsize,anchor=south east] at (0.315,0.06) {$0.5$}; 
            \node[font=\scriptsize,anchor=south east] at (0.443,0.06) {$1$}; 
            \node[font=\scriptsize,anchor=south east] at (0.625,0.06) {$0$};
            \node[font=\scriptsize,anchor=south east] at (0.78,0.06) {$0.5$};
            \node[font=\scriptsize,anchor=south east] at (0.925,0.06) {$1$}; 
            \node[font=\scriptsize,anchor=south east] at (0.625,0.52) {$0$};
            \node[font=\scriptsize,anchor=south east] at (0.78,0.52) {$0.5$};
            \node[font=\scriptsize,anchor=south east] at (0.925,0.52) {$1$};
            \node[font=\scriptsize,anchor=south east] at (0.13,0.555) {$-30$}; 
            \node[font=\scriptsize,anchor=south east] at (0.13,0.702) {$0$}; 
            \node[font=\scriptsize,anchor=south east] at (0.13,0.845) {$30$};
            \node[font=\scriptsize,anchor=south east] at (0.13,0.09) {$2.5$}; 
            \node[font=\scriptsize,anchor=south east] at (0.13,0.27) {$10$}; 
            \node[font=\scriptsize,anchor=south east] at (0.13,0.39) {$15$};
            \node[font=\scriptsize,anchor=south east] at (0.59,0.555) {$-20$}; 
            \node[font=\scriptsize,anchor=south east] at (0.59,0.702) {$0$}; 
            \node[font=\scriptsize,anchor=south east] at (0.59,0.845) {$20$};
            \node[font=\scriptsize,anchor=south east] at (0.59,0.09) {$2.5$}; 
            \node[font=\scriptsize,anchor=south east] at (0.59,0.15) {$5$}; 
            \node[font=\scriptsize,anchor=south east] at (0.59,0.39) {$15$};
        \end{scope}
    \end{tikzpicture}
\end{minipage}%
\hfill 
\begin{minipage}{.5\textwidth}
    \centering
    \begin{tikzpicture}
        \node[anchor=south west,inner sep=0] (image) at (0,0) {\includegraphics[width=\linewidth]{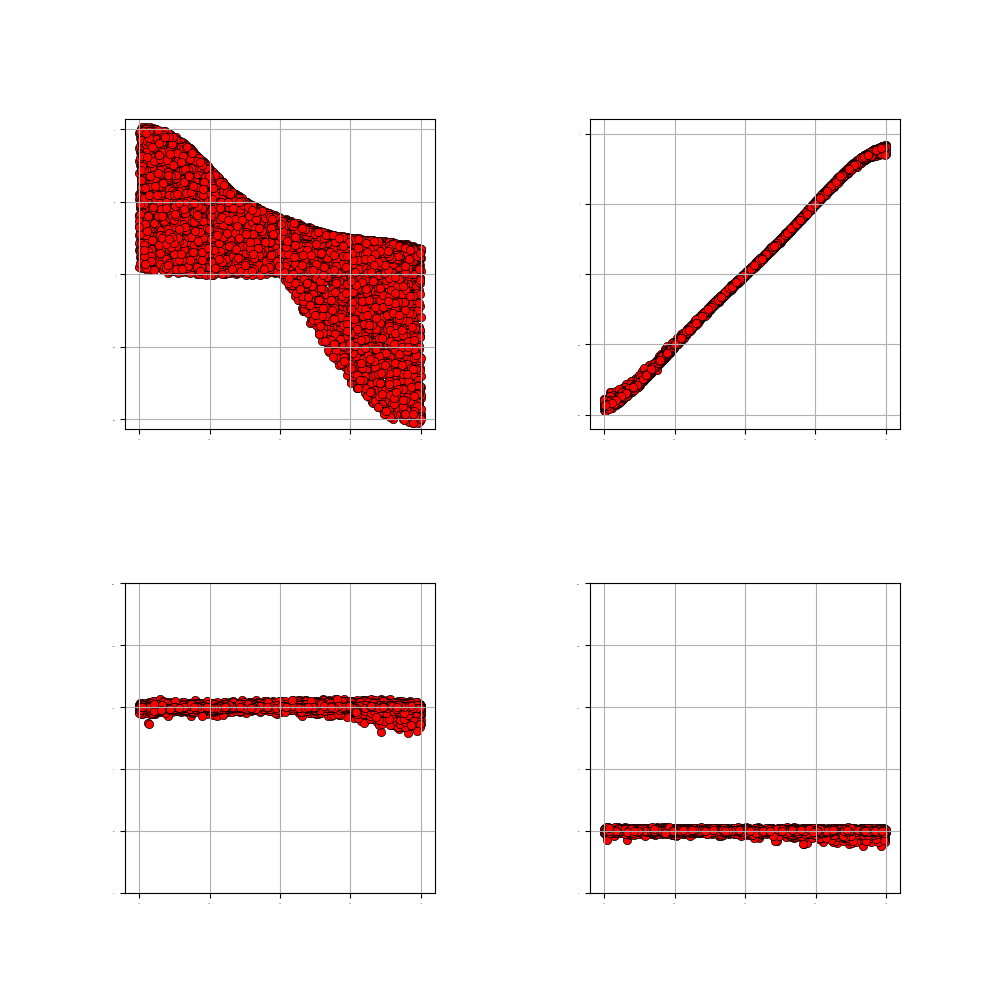}};
        \begin{scope}[x={(image.south east)},y={(image.north west)}]
            \node[font=\large,anchor=south east] at (0.57,0.9) {\ginn};
            \node[font=\small,anchor=south east ,rotate=90] at (0.08,0.765) {\(\netsol_{x_1}\)}; 
            \node[font=\small,anchor=south east ,rotate=90] at (0.55,0.765) {\(\netsol_{x_3}\)};
            \node[font=\small,anchor=south east ,rotate=90] at (0.08,0.31) {\(\netsol_{x_4}\)};
            \node[font=\small,anchor=south east ,rotate=90] at (0.55,0.31) {\(\netsol_{x_5}\)};
            \node[font=\small,anchor=south east] at (0.315,0.475) {\(x_1\)};
            \node[font=\small,anchor=south east] at (0.78,0.475) {\(x_3\)};
            \node[font=\small,anchor=south east] at (0.315,0.02) {\(x_4\)}; 
            \node[font=\small,anchor=south east] at (0.78,0.02) {\(x_5\)};
            \node[font=\scriptsize,anchor=south east] at (0.162,0.52) {$0$}; 
            \node[font=\scriptsize,anchor=south east] at (0.315,0.52) {$0.5$}; 
            \node[font=\scriptsize,anchor=south east] at (0.443,0.52) {$1$};
            \node[font=\scriptsize,anchor=south east] at (0.162,0.06) {$0$}; 
            \node[font=\scriptsize,anchor=south east] at (0.315,0.06) {$0.5$}; 
            \node[font=\scriptsize,anchor=south east] at (0.443,0.06) {$1$}; 
            \node[font=\scriptsize,anchor=south east] at (0.625,0.06) {$0$};
            \node[font=\scriptsize,anchor=south east] at (0.78,0.06) {$0.5$};
            \node[font=\scriptsize,anchor=south east] at (0.925,0.06) {$1$}; 
            \node[font=\scriptsize,anchor=south east] at (0.625,0.52) {$0$};
            \node[font=\scriptsize,anchor=south east] at (0.78,0.52) {$0.5$};
            \node[font=\scriptsize,anchor=south east] at (0.925,0.52) {$1$};
            \node[font=\scriptsize,anchor=south east] at (0.13,0.555) {$-30$}; 
            \node[font=\scriptsize,anchor=south east] at (0.13,0.702) {$0$}; 
            \node[font=\scriptsize,anchor=south east] at (0.13,0.845) {$30$};
            \node[font=\scriptsize,anchor=south east] at (0.13,0.09) {$2.5$}; 
            \node[font=\scriptsize,anchor=south east] at (0.13,0.27) {$10$}; 
            \node[font=\scriptsize,anchor=south east] at (0.13,0.39) {$15$};
            \node[font=\scriptsize,anchor=south east] at (0.59,0.555) {$-20$}; 
            \node[font=\scriptsize,anchor=south east] at (0.59,0.702) {$0$}; 
            \node[font=\scriptsize,anchor=south east] at (0.59,0.845) {$20$};
            \node[font=\scriptsize,anchor=south east] at (0.59,0.09) {$2.5$}; 
            \node[font=\scriptsize,anchor=south east] at (0.59,0.15) {$5$}; 
            \node[font=\scriptsize,anchor=south east] at (0.59,0.39) {$15$};
        \end{scope}
    \end{tikzpicture}
\end{minipage}
\caption{\friedman. \textit{Left two columns}: \snn~with \(n=200\) ($\rmse_U = 0.51$). \textit{Right two columns}: \ginn~with \(n=200\) and $m=1000$ ($\rmse_U = 0.04$). The plot shows the learned gradient $U_{x_i}$ for each dimension $x_i, i=1, \dots, 5$ and point \(\vecx_n \in \dataTp_{\acro{test}}\) ($U_{x_2}$ is not reported as similar to $U_{x_1}$).}
\label{fig:Friedman_U_grad_tot}
\vskip -0.2in

\end{figure*}

See Figure \ref{fig:Friedman_U_grad_tot} for additional results.

\begin{table*}[h]
\caption{\friedman. $\rmse_\partial$ for different values of $\ndataTg$ and across input dimensions when $\sigma^2=0$.
}
\label{RMSE_NN_grad_LT_l1_l2}
\vskip 0.15in
\begin{center}
\begin{small}
\begin{tabular}{lcccccccccc}
\toprule
&  \multicolumn{5}{c}{\snn with $\ell_1$} & \multicolumn{5}{c}{\snn with $\ell_2$} \\
\cmidrule(lr){2-6} 
\cmidrule(lr){7-11}
\(\ndataTg\) & \(\netsol_{x_1}\) & \(\netsol_{x_2}\) & \(\netsol_{x_3}\) &  \(\netsol_{x_4}\) &  \(\netsol_{x_5}\) & \(\netsol_{x_1}\) &\(\netsol_{x_2}\) & \(\netsol_{x_3}\) & \(\netsol_{x_4}\) & \(\netsol_{x_5}\) \\
\cmidrule(lr){1-1} 
\cmidrule(lr){2-6}
\cmidrule(lr){7-11}
50    & 6.1 & 6.5 & 8 & 3.0 & 2.6  & 5.9 & 6.0 & 7.9 & 2.9 & 2.28 \\
100   & 3.9  & 3.7 & 2.8  & 1.27  & 0.9  & 3.2 & 3.1 & 2.0 & 1.1 & 0.67 \\
200   & 1.1 & 1.0 & 1.1 & 0.5 & 0.3  & 1.3 & 1.1 & 0.85 & 0.44 & 0.25 \\
500   & 0.6 & 0.63 & 0.40  & 0.22  & 0.07  & 0.59 & 0.58 & 0.52 & 0.21 & 0.11 \\
\bottomrule
\end{tabular}
\end{small}
\end{center}
\vskip -0.1in
\end{table*}

\begin{table}[ht]
\caption{\friedman. $\rmse_U$ for different values of $c$ with $\ndataTg=200$ and $\ndataCg=1000$ for \ginn. For each value of $c$ the lowest \rmse is bolded.}
\label{RMSE_noise_tab}
\begin{center}
\begin{small}
\begin{tabular}{lccccc}
\toprule
 & \(c=0.0\) & \(c=0.01\) & \(c=0.03\) & \(c=0.05\) \\
\midrule
\snn  & 0.51 & 0.53 & 0.66 & 0.75\\
\snn~- \(\ell_1\)  & 0.13 & 0.18 & 0.27 & 0.45\\
\snn~- \(\ell_2\)  & 0.13 & 0.18 & 0.25 & 0.40\\
\ginn & \textbf{0.04} & \textbf{0.07} & \textbf{0.19} & \textbf{0.30}\\
\acro{st} & 0.17 & 0.18 & 0.24 & 0.31\\
\bottomrule
\end{tabular}
\end{small}
\end{center}
\vskip -0.in
\end{table}

\section{Additional results for \burgers}
\label{Burger_add_exp}
\begin{figure}[h]
\begin{center}
\includegraphics[width=0.8\columnwidth]{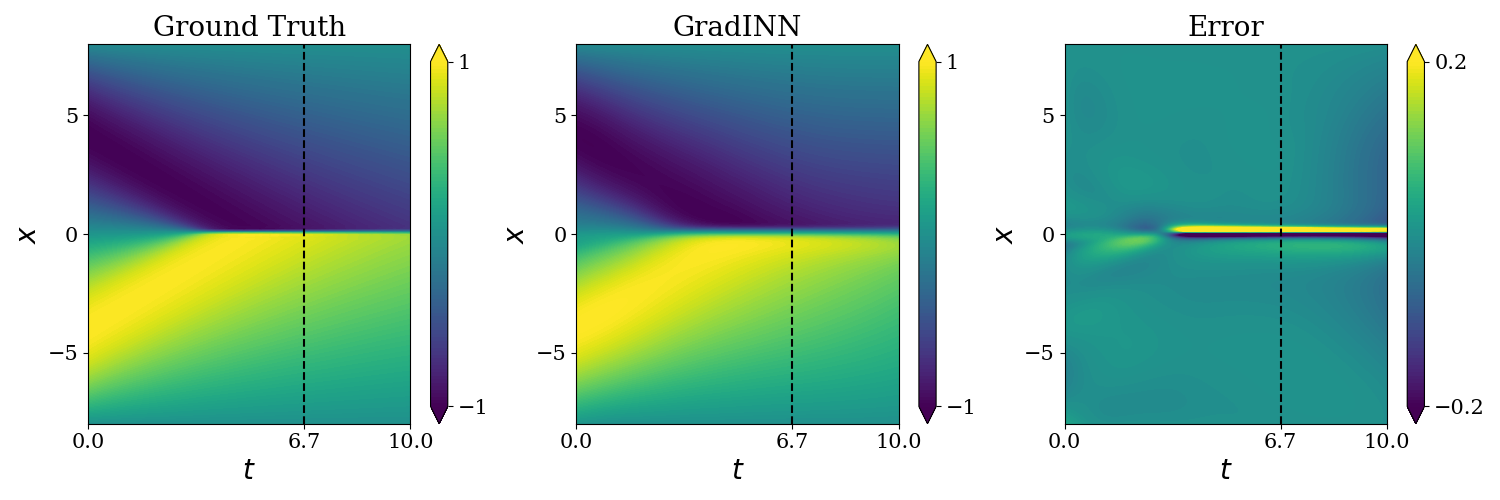}
\caption{\burgers with $\nu=0.01/\pi$. \textit{Left}: \ginn~predicted solution. \textit{Middle:} Ground truth Solution. \textit{Right:} Difference between \ginn~and Ground truth.}
\label{fig:Burgers_lowNU}
\end{center}
\end{figure}

To further analyze the phenomenon of solution oversmoothing shown in Fig.\ref{burgers_fig}, we repeated the \burgers experiment considering the same initial condition $u(0, x) = -\sin(\pi x/8)$ but $\nu=0.01/\pi$, a condition that leads to steeper gradients. We observe that, under these settings, the oversmoothing of the predicted solution is more pronounced (see Fig. \ref{fig:Burgers_lowNU}) thus shedding light on one of \ginn's limitation when using a smooth prior belief to approximate a function with very steep local derivatives. This emphasizes the importance of developing methods that allows encoding more complex prior beliefs in $\netauxg$ so as to better predict systems with discontinuities or high local gradients.


\section*{Aknowledgements}
F. D. aknowledges that this study was carried out within the FAIR-Future Artificial Intelligence Research and received funding from the European Union Next-GenerationEU (PIANO NAZIONALE DI RIPRESA E RESILIENZA (PNRR)–MISSIONE 4 COMPONENTE 2, INVESTIMENTO 1.3---D.D. 1555 11/10/2022, PE00000013). This manuscript reflects only the authors’ views and opinions; neither the European Union nor the European Commission can be considered responsible for them.

\bibliography{Bibliography}


\end{document}